\documentclass[10pt,twocolumn,twoside]{IEEEtran} \usepackage{graphicx} 
\usepackage{lipsum}
\usepackage{amsmath}
\usepackage{amsfonts}
\usepackage{amssymb}
\usepackage{bm}
\usepackage{tikz}
\usepackage{pgfplots}
\usepackage{hyperref}
\usepackage{cleveref}
\usepackage{comment}
\usetikzlibrary{calc,shapes.geometric}

\usepackage{amsthm}
\usepackage{xintexpr}
\usepackage{xinttools} 
\usepackage{tikz}
\usetikzlibrary{calc}

\def\biglen{20cm} 
\tikzset{
  half plane/.style={ to path={
       ($(\tikztostart)!.5!(\tikztotarget)!#1!(\tikztotarget)!\biglen!90:(\tikztotarget)$)
    -- ($(\tikztostart)!.5!(\tikztotarget)!#1!(\tikztotarget)!\biglen!-90:(\tikztotarget)$)
    -- ([turn]0,2*\biglen) -- ([turn]0,2*\biglen) -- cycle}},
  half plane/.default={1pt}
}

\def\maxxy{4} 

\usepackage[citestyle=numeric-comp,sorting=none,url=false,doi=false,isbn=false]{biblatex}
\AtEveryBibitem{\clearfield{month}}
\AtEveryBibitem{\clearfield{day}}
\AtEveryBibitem{\clearfield{pages}}

\newcommand{\R}{\mathbb{R}}

\newcommand{\N}{\mathbb{N}}
\newcommand{\e}{\mathrm{e}}

\DeclareMathOperator*{\argmin}{arg\,min}

\title{Optimal Multi-Robot Communication-Aware Trajectory Planning by Constraining the Fiedler Value}
\author{Jeppe Heini Mikkelsen, Roberto Galeazzi, and Matteo Fumagalli \\ Technical University of Denmark, Automation and Control Group}
\date{November 2023}

\begin{document}

\maketitle

\begin{abstract}
    The paper present a novel approach for the solution of the Multi-Robot Communication-Aware Trajectory Planning, which builds on a general optimisation framework where the changes in robots positions are used as decision variable, and linear constraints on the trajectories of the robots are introduced to ensure communication performance and collision avoidance. The Fiedler value is adopted as communication performance metric. The validity of the method in computing both feasible and optimal trajectories for the robots is demonstrated in simulation. Results show that the constraint on the Fiedler value ensures that the robot network fulfils its objective while maintaining communication connectivity at all times. Further, the paper shows that the introduction of approximations for the constraints enables a significant improvement in the computational time of the solution, which remain very close to the optimal solution.
\end{abstract}
\section{Introduction}
Multi-robot systems have received a lot of attention in recent years, especially in information gathering scenarios such as exploration \cite{Amigoni2017MultirobotSurvey} and search-and-rescue \cite{Queralta2020CollaborativeVision}. One of the main limitations in multi-robot systems is communication. Almost all mobile robots rely on some form of wireless communication to transmit information between each other and/or to a ground station. For multi-robot systems, inter-robot communication is often necessary for coordination, e.g., in consensus \cite{Olfati-Saber2007ConsensusSystems} and task allocation \cite{Quinton2023MarketSurvey}. There is also often a need for an operator to monitor the robots to ensure nominal behaviour and intervene if needed, necessitating communication with a base station. Wireless communication technologies have a number of limitations, chief among them how far and at which rate they can transmit data. These limitations are mainly due to three physical effects: path loss due to signal attenuation, shadowing due to obstacles, and multi-path fading due to reflections \cite{Muralidharan2021Communication-AwareCommunication}. This produces a coupling between the robots position and its ability to communicate, necessitating multi-robot communication-aware trajectory planners (MR-CaTP).
\begin{figure}[ht]
    \centering
    \begin{tikzpicture}[scale=0.75]

        \draw[ultra thick] (2.29289321881,-1.70710678119) arc (-120:30:1) -- cycle;
        \draw[ultra thick] (3,-1) -- (2.5,-0.5);
        \filldraw (2.5,-0.5) circle (2pt);
        \draw[ultra thick, red] (2.4510,-0.2549) -- (2.1471,1.2646);
        \draw[ultra thick, red] (0.7354,-0.1471) -- (2.2549,-0.4510);
        \draw[ultra thick, red] (0.5303,0.5303) -- (1.4697,1.4697);
        \draw[ultra thick, red] (-1.2500,2.0000) -- (1.2500,2.0000);
        \draw[ultra thick, red] (-2.2785,1.3036) -- (-2.7215,0.1964);
        \draw[ultra thick, red] (-0.7398,-0.1233) -- (-2.2602,-0.3767);
        \draw[ultra thick, red] (-1.4697,1.4697) -- (-0.5303,0.5303);
        \draw[ultra thick, red] (-4.2602,1.6233) -- (-2.7398,1.8767);
        \draw[ultra thick, red] (-4.4697,0.9697) -- (-3.5303,0.0303);
    
        \foreach \x/\y/\scale/\rotate in {0/0/0.25/30, -2/2/0.25/30, 2/2/0.25/-45, -3/-0.5/0.25/90, -5/1.5/0.25/60}
            {
                \begin{scope}[shift={(\x,\y)}, scale=\scale, rotate=\rotate]
                    \draw[ultra thick] (-1,0) -- (1,0);
                    \draw[ultra thick] (0,-1) -- (0,1);
                    \draw[ultra thick] (-1.5,0) circle (0.5);
                    \draw[ultra thick] (1.5,0) circle (0.5);
                    \draw[ultra thick] (0,-1.5) circle (0.5);
                    \draw[ultra thick] (0,1.5) circle (0.5);
                \end{scope}
            }

            \draw[ultra thick, blue, dashed] (-1.25,-0.5) arc (-180:-270:3.5);

    \end{tikzpicture}
    \caption{An example of an ad-hoc network where five drones are communicating wirelessly with each other and a ground station antenna. Three of the drones are outside the communication range of the antenna (blue dashed line), and therefore have to rely on the other drones relaying their information.}
    \label{fig:network}
\end{figure}
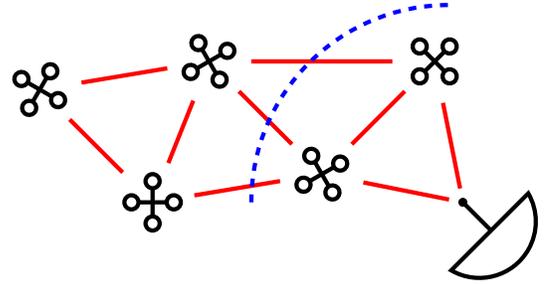
There are many different technologies and protocols that can be used for wireless communication in robotics. This paper focuses on wireless ad-hoc networks. Ad-hoc networks are not reliant on the availability of pre-existing communication infrastructure such as routers or access-points, instead all nodes in the network participate in the routing of data, though pre-existing infrastructure can exist in an ad-hoc setting, see \cref{fig:network}. This enables the network to reach further than what a single centralised routing station can achieve, and also adds robustness since there is no reliance on a centralised node to route the data. It does, however, complicate the planning in that the communication not only becomes coupled with the position of the robots themselves, but also the positions of the robots relative to each other, making the complexity of ensuring communication grow exponentially with the number of robots. To develop planners and controllers that ensure communication, communication must first be modelled as a function of the robot states. There are a range of methods to do this. Some model the physical parameters of the network, some model the link quality between communication devices, while others use proximity-based models.

The received signal strength indicator (RSSI) quantifies the physical power present in a signal and the signal-to-noise ratio (SNR) how much of the signal is discernible by the receiver. In \cite{Hsieh2006TowardsGuarantees} a reactive stop and return strategy is proposed, based on SNR measurements. In \cite{Lindhe2009UsingFading} three stopping strategies, based on RSSI measurements, are proposed, where each robot loads sensor data into a buffer and routinely offloads it via communication. In \cite{Dixon2009MaintainingControl} an extremum seeking controller is proposed to maximise the SNR in a chain. In \cite{Daniel2010ASwarms} the authors present a method for maximising coverage of a swarm while ensuring communication, using RSSI measurements within clusters of robots. In \cite{Tardioli2010EnforcingMissions}, links are preserved using virtual spring-dampers between robots if the RSSI falls below a threshold. These methods measure the RSSI or SNR and react on it immediately. Reactive approaches have the risk of leading to locally optimal solutions that the robots cannot escape. To have less reactive approaches that can plan longer into the future, the effect of the environment on the RSSI must be modelled. In \cite{Marchukov2016Multi-robotConstraints,Marchukov2017Communication-awareDeployment} the authors present methods for building a communication tree to visit goals, using virtual goals to ensure communication. Both methods model the RSSI with path loss and shadowing. In \cite{Woosley2020Multi-robotConstraints} the authors present a method for collecting information while ensuring periodic communication, based on collecting measurements of information and communication quality. Modelling RSSI and SNR with shadowing and multipath fading is computationally demanding, with a dependence on high-fidelity maps of the environment with known material properties. Furthermore, due to the complexity of signal propagation, substantial discrepancies between modelled and actual RSSI or SNR can occur, potentially leading to dire consequences. Measuring the RSSI directly is possible, since many WiFi adapters can do that, but cannot be used to reliably predict it over longer trajectories. 

Link quality indicates how well a connection between two nodes have been established, often quantified by the probability of a successful communication packet transmission. In \cite{Zavlanos2013NetworkNetworks} the authors propose a two-stage approach that iteratively computes the optimal routing variables and finds velocities that ensure communication. Assuming static robots, network integrity constraints are constructed from the link reliabilities, and the optimal routing variables are found as the solution to a distributed convex optimisation problem. A potential field is then constructed such that positions where network integrity is compromised become virtual obstacles. Instead of using a potential field, the method in \cite{Kantaros2016GlobalEnvironments} computes a reference position obeying a linearised version of the network integrity constraint. A communication tree for servicing tasks is then formed by distributively electing teams to expand the tree toward task locations. In \cite{Stephan2016HybridSystems,Stephan2017ConcurrentSystems} a centralised planner generates a sequence of waypoints obeying a stochastic version of the network integrity constraint, allowing robots to remain reactive while avoiding local minima. In \cite{Mox2020MobileDemand}, robots are divided into a network team, responsible for ensuring communication, and a task team, which assumes communication is provided during task execution, enhancing versatility. Communication networks can also be modelled as graphs, consisting of vertices or nodes, representing devices, and edges, representing communication links. By associating link quality with the corresponding edge, graph-theoretic concepts can be used to ensure communication. A common approach is to calculate the Fiedler value, which is the second smallest eigenvalue of the graphs Laplacian matrix. This value is positively correlated with the connectivity of the network, and as long as it stays strictly positive, the network is connected. In \cite{DeGennaro2006DecentralizedSystems}, the Fiedler value is maximised by distributively calculating its gradient with respect to the Laplacian. Robots then use a potential field to realise the desired Laplacian. In \cite{Kim2006OnLaplacian} the Fiedler value is maximised using semi-definite programming, while ensuring collision avoidance using a linearised distance constraint. In \cite{Zavlanos2007DistributedNetworks} the Laplacian is used to construct a potential field that grows as the Fiedler value goes to zero and as the robots get closer. In \cite{Yang2008DecentralizedNetworks} the Fiedler value and its corresponding eigenvector, which the robots use to calculate its gradient to move along to maximise communication, is calculated distributively. In \cite{EthanStump2008ConnectivityTeams}, the Fiedler value for a communication chain is maximised, with the constraint of maintaining k-connectivity to ensure communication between its head and tail. In \cite{DeSanBernabeClemente2014WirelessRobots} an ad-hoc network is repaired by measuring the link quality at different locations, and adding relay robots to ensure connectivity and redundancy. In \cite{Mostofi2005CommunicationApproach}, the authors propose an information-based approach for communication in target tracking by modelling link quality as additive white noise, with the robots optimising their placement to maximise information on the target.

Many methods opt for a proximity-based approach, where a communication link between two devices is assumed to exist if they are within communication range. A common approach for ensuring preservation of proximity between robots is by using a potential field that grows as the distance between robots approaches the maximum communication range. In \cite{Esposito2006MaintainingObstacles} the authors present a potential field for preserving proximity while also preserving line-of-sight (LOS) to mitigate shadowing. In \cite{Su2010RendezvousConnectivity} an attractive potential field is used to achieve rendezvous, where a sudden large attraction when robots enter each others proximity is avoided by using hysteresis when adding connections. In \cite{Li2013AObstacles} the authors use a potential field to preserve connections and avoid collisions in a fixed network topology. In \cite{Sabattini2013DistributedMaintenance} a decentralised estimate of the Fiedler value is used to construct a potential field that grows as the Fiedler value gets closer to a lower bound. In \cite{Mondal2018TrajectoryAssurance} the authors present a bounded potential function that preserve communication and provides collision avoidance. In \cite{Ji2007DistributedConnectedness} a weighted consensus protocol is used to rendezvous while maintaining communication, where the weights are calculated as edge-tensions that grow as the distance between the robots approaches the maximum communication range. In \cite{Sabattini2015DecentralizedGraphs} the authors use edge tension to ensure communication in directed graphs, but only when the Fiedler value is below some threshold. In \cite{Jaleel2018ANetworks} the edges that must be preserved for connectivity is found, and weighted consensus with edge tension is used to rendezvous. In \cite{Han2010AutonomousLeaders} the authors propose a method for navigating a swarm according to some covert leaders. The proximity around each robot is divided into zones to ensure collision avoidance, alignment, and proximity. Sometimes communication links need to be removed, where the challenge is to do so without disjoining the network. In \cite{Zavlanos2007DistributedNetworks,Michael2009MaintainingNetworks} the authors propose distributed methods for adding and deleting communication links while preserving communication. In \cite{Griparic2022Consensus-BasedSystems} the authors propose a distributed method for adding and deleting communication links to track a desired Fiedler value in static networks. In \cite{Spanos2004RobustVehicles} a metric for network robustness is proposed. In \cite{Zavlanos2005ControllingGraphs} the authors present a method for constructing a linear constraint on the velocities of the robots to ensure communication using k-connectivity. In \cite{Spanos2005MotionConstraints} a method for transitioning from one connected configuration to another while preserving the robustness of the network is presented. In \cite{Schouwenaars2006Multi-vehicleCommunication} the authors present a mixed-integer programme to ensure line-of-sight communication in a chain while avoiding obstacle collisions. In \cite{Rooker2007Multi-robotNetworking} the authors present a sampling based planner for ensuring communication. In \cite{Hollinger2012MultirobotExperiments} the authors present a method for ensuring periodic connectivity by sequentially planning robot paths, with each robot being connected to the others at the end of the planning period. In \cite{Shi2021Communication-AwareMaximization} the robots compute their desired positions, form a minimum spanning tree, and compute the optimal deviation that preserve communication.

In summary, three distinct approaches to modelling communication in robotics have been presented. The first and most realistic approach use the signal propagation through the environment to ensure that there is sufficient signal strength for the robots to communicate. However, these methods are either reactive or they assume full knowledge of the environment, including the material properties of the obstacles, which is rarely known. The second approach models the link quality, which is the probability of a communication packet transmitted between two devices being received. It is an abstraction on the outcome of the aforementioned signal propagation and is therefore more simplistic. Lastly, the simplest approaches uses proximity based methods, where the existence of a communication link between two devices is assumed when they are within a certain range of each other. These methods are severely limited since the quality of the connection is assumed constant when they are within range. In this paper we have opted for a link quality approach, as we believe it is a fair compromise between fidelity and abstraction.

In this paper, the Fiedler value is used as a communication performance metric. The main contribution of this paper is the reformulation of a generic optimal MR-CaTP problem into a problem that is more tractable, by imposing linear constraints on the robot trajectory to ensure communication performance and collision avoidance. MR-CaTP based on the Fiedler value is a well established area of research. However, most of the methods are tailored to a specific use-case and do not provide hard constraints on the Fiedler value. The benefit of the method proposed in this paper is its versatility, since it is applied in a general optimisation framework where the change in robot positions are the decision variables. We also demonstrate that it can be used in non-optimal control and planning, by computing the smallest deviation to the desired robot inputs that preserve communication. As mentioned, there is a rich literature on communication-aware planning algorithms based on the Fiedler value and its gradient, but to our knowledge none have formulated a linear constraint using the Fiedler value in a general optimisation framework. The code for the simulation results in this paper is provided at \url{https://github.com/JeppHMikk/MR-CaTP}.

Throughout the paper, we adopt the following notation style: italic symbols $x/X$ denote scalars; bold italic symbols \(\bm{x}/\bm{X}\) denote vectors; bold non-italic symbols \(\mathbf{x}/\mathbf{X}\) denote matrices; and calligraphic symbols \(\mathcal{X}\) denote sets. Notations \(\R_{\geq0}\) and \(\R_{>0}\) denote non-negative and strictly positive real numbers, respectively. $\N_N=\{1,\dots,N\}$ denotes the natural numbers up to $N$. Square brackets $[\bm{x},\bm{y}]$ indicate column concatenation and parentheses $(\bm{x},\bm{y})$ indicate row concatenation. Unless explicitly stated, all vectors are column vectors.

\section{Robot and Communication Network Model}
Consider a swarm of $N$ robots, $\mathcal{V} = \N_N$, where the position of robot $i$ at time $t$ is denoted as $\bm{p_i}(t)\in\R^n$. The robots are assumed to be holonomic and have controllers capable of tracking a reference position. Furthermore, the physical space that each robot takes up is represented as a minimum circumscribing hyper-sphere with centre at $\bm{p_i}(t)$ and radius $r_i$. The robots are modelled as moving in discrete time steps $\Delta t$
\begin{equation}\label{eq:discrete_model}
    \bm{p^{k+1}} = \bm{p^k} + \bm{u^k},
\end{equation}
where the superscript $k$ denotes the k\textsuperscript{th} discrete time index
\begin{equation}
    \bm{p_i^k} = \bm{p_i}(k\Delta t), \quad \bm{u_i^k} = \bm{u_i}(k\Delta t),
\end{equation}
and $\bm{u_i^k}$ is the change in position of robot $i$ between time-steps $k$ and $k+1$, henceforth referred to as the input to robot $i$ at time-step $k$.
The concatenated robot positions and inputs are defined as $\bm{p^k} = (\bm{p_1^{k}},\bm{p_2^{k}},\dots,\bm{p_N^{k}}) \in \R^{Nn}$ and $\bm{u^k} = (\bm{u_1^{k}},\bm{u_2^{k}},\dots,\bm{u_N^{k}}) \in \R^{Nn}$ respectively.
The robots exchange information with each other through wireless communication. The communication network can be modelled as a time-dependent undirected graph $\mathcal{G}(t) = (\mathcal{V},\mathcal{E}(t))$ where $\mathcal{E}(t) \in \mathcal{V}\times\mathcal{V}$ is the edge set representing time-varying communication links between robot pairs. Furthermore, with each edge there is an associated weight $w_{ij} \in \{\mathcal{W} : \mathcal{E}\times\R^2\times\R^2 \rightarrow [0,1]$\}, where $w_{ij}$ is the weight associated with the edge from vertex $i$ to $j$. 
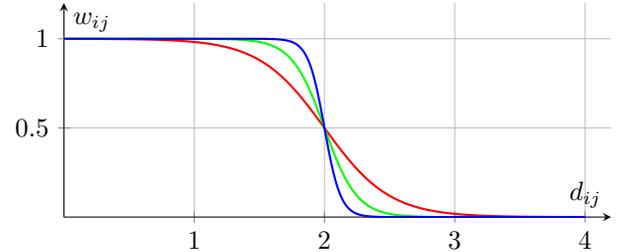
\begin{figure}[ht]
    \centering
    \begin{tikzpicture}
        \begin{axis}[
            domain=0:4, 
            samples=500,
            axis lines=middle, 
            grid=both, 
            xmin=0, 
            xmax=4.2, 
            ymin=0, 
            ymax=1.2, 
            xlabel={$d_{ij}$}, 
            ylabel={$w_{ij}$},
            width=\linewidth,
            height=0.5\linewidth
        ]
            \addplot[red, thick] {1 - 1/(1 + exp(-4*(x - 2)))};
            \addplot[green, thick] {1 - 1/(1 + exp(-8*(x - 2)))};
            \addplot[blue, thick] {1 - 1/(1 + exp(-16*(x - 2)))};
            \node at (axis cs:2,-0.1) [anchor=north] {$d_0$};
            \node at (axis cs:0,-0.1) [anchor=north] {$0$};
            \node at (axis cs:0,1) [anchor=east] {$1$};
        \end{axis}
    \end{tikzpicture}
    \caption{Link quality with increasing values of $\alpha$, in order of red, green, and blue.}
    \label{fig:packet_arrival}
\end{figure}
The edge weight represents the link quality, i.e., the probability that a communication packet transmitted by robot $i$ is received successfully by robot $j$. The link quality is dependent on the distance between the two robots, and is modelled using the logistic function \cite{Zavlanos2011Graph-theoreticNetworks}
\begin{equation}
    w_{ij}(t) =
    \begin{cases}
        \dfrac{\e^{-\alpha(d_{ij}(t) - d_{50})}}{1 + \e^{-\alpha(d_{ij}(t) - d_{50})}} \in [0,1] \ \forall \ (i,j) \in \mathcal{E}, \\
        0 \quad \text{else},
    \end{cases}
\end{equation}
\begin{equation}
    d_{ij}(t) = ||\bm{p_i}(t) - \bm{p_j}(t)||_2,
\end{equation}
where $d_{50}$ is the distance at which the link quality is $50\%$ and $\alpha$ is $4$ times how fast it attenuates at this distance. When the distance is significantly lower than $d_{50}$ the link quality is close to $1$, and when it is significantly higher the link quality is close to $0$, see \cref{fig:packet_arrival}. Since $d_{ij}(t) = d_{ji}(t)$, it follows that $w_{ij}(t) = w_{ji}(t)$. This function is used to approximate the link quality due to it being monotonically decreasing, since it is expected that the link quality only deteriorates as the distance grows, and due to it being $\mathcal{C}^1$ continuous, which will be elaborated on later. The specific values of $d_{50}$ and $\alpha$ depend on the communication hardware and must be determined experimentally. In \cite{DeSanBernabeClemente2014WirelessRobots} the authors establish a sigmoidal function that relates distance to link quality for some actual robots.

\section{Communication Performance Metric}
To ensure communication performance, a metric has to be determined. 
For this paper, it is decided that the metric both has to capture the condition that the network has to remain connected and to which degree it is. The Fiedler value meets both of these criteria. Given a graph $\mathcal{G}(t)$, the adjacency matrix $\mathbf{A}(t) = (a_{ij}(t)) \in [0,1]^{N\times N}$ is a hollow symmetric matrix and the degree matrix $\mathbf{D}(t) = (d_{ij}(t)) \in [0,N-1]^{N\times N}$ is a diagonal matrix, where each element is defined as 
\begin{gather}
    a_{ij}(t) = w_{ij}(t) \ \forall \ i \neq j \in \mathcal{V}\times\mathcal{V}, \\ d_{ii}(t) = \sum_{j\in\mathcal{V}} a_{ij}(t) \ \forall \ i \in \mathcal{V}.
\end{gather}
The graph Laplacian is defined as
\begin{equation}
    \mathbf{L}(t) = \mathbf{D}(t) - \mathbf{A}(t),
\end{equation}
with its eigenvalues and associated eigenvectors being defined as $\lambda_1(t) \leq \lambda_2(t) \leq \dots \leq \lambda_N(t)$ and $\bm{v_1}(t),\bm{v_2}(t),\dots,\bm{v_N}(t)$ respectively. It can be shown that the eigenvalues of $\mathbf{L}(t)$ are bounded between $0$ and $N$, with the smallest eigenvalue and corresponding eigenvector being $\lambda_1(t) = 0 \ \forall \ t\in\R$ and $\bm{v_1}(t) = \bm{1} \ \forall \ t\in\R$. The second smallest eigenvalue $\lambda_2(t)$ is known as the Fiedler value. If $\lambda_2(t) = 0$ the graph is disjoint at time $t$, i.e., there is a division of the vertex set such that all the weights between vertices across the sets are zero
\begin{equation}
    \begin{split}
        \lambda_2(t) = 0 \Leftrightarrow \exists \ \mathcal{V}_1, \mathcal{V}_2 \ | \ w_{ij}(t) = 0 \ \forall \ i,j \in \mathcal{V}_1 \times \mathcal{V}_2 \dots \\ \land \ ( \mathcal{V}_1 \cap \mathcal{V}_2 = \emptyset ) \ \land \ ( \mathcal{V}_1 \cup \mathcal{V}_2 = \mathcal{V} ),
    \end{split}
\end{equation}
which means that there are parts of the network that are unable to communicate. If $\lambda_2(t) = N$ the graph is complete
\begin{equation}
    \lambda_2(t) = N \Leftrightarrow w_{ij}(t) = 1 \ \forall \ i \neq j \in \mathcal{V} \times \mathcal{V},
\end{equation}
which means that all node pairs can communicate with $100\%$ packet reception rate \cite{DeGennaro2006DecentralizedSystems}. 
Many multi-agent algorithms rely on consensus. In consensus algorithms devices in a network reach agreement on variables by exchanging information. It can be shown that consensus dynamics converge asymptotically with time constant $\tau = \lambda_2^{-1}$ \cite{Mesbahi2010GraphNetworks}. Furthermore, another property of the Fiedler value is that it lower bounds the isoperimetric number, also known as the Cheeger number \cite{Mesbahi2010GraphNetworks}. The Cheeger number measures the size of the smallest bottleneck in the network and can therefore be used to represent connectivity, as the rate at which data can be pushed through the bottleneck represents an upper bound on the effective bandwidth of the network.

Therefore, the Fiedler value is also a metric of how connected the graph is and can therefore be used as a performance metric for the communication network. Since the edge weights in the graph are only approaching zero in the limit
\begin{equation}
    \lim_{d_{ij}(t) \rightarrow \infty} w_{ij}(t) = 0 \ \forall \ i \neq j \in \mathcal{V}\times\mathcal{V},
\end{equation}
the Fiedler value will also only become zero in the limit
\begin{equation}
    \lim_{d_{ij}(t) \rightarrow \infty \forall (i,j)\in\mathcal{E}}\lambda_2(t)=0.
\end{equation}
Therefore, the Fiedler value needs to remain greater than or equal to a lower bound $\lambda_2(t) \geq \underline{\lambda}_2 > 0 \ \forall \ t \in \R$, where $\underline{\lambda}_2$ is greater than $0$ to ensure that the network never becomes "barely" connected. The lower bound $\underline{\lambda}_2$ is set by the operator and can be based on, e.g., the slowest allowed convergence time for a consensus algorithm. For algorithmic purposes, the graph, its adjacency, degree, and Laplacian matrix, as well as its Fiedler value, is calculated at discrete time steps henceforth denoted as
\begin{equation}
    \begin{aligned}
        &\mathbf{A}(k\Delta t) = \mathbf{A^k}, &\mathbf{D}(k\Delta t) = \mathbf{D^k} \\
        &\mathbf{L}(k\Delta t) = \mathbf{L^k}, &\lambda_2(k\Delta t) = \lambda_2^k.
    \end{aligned}
\end{equation}

\section{Optimisation Problem}\label{sec:optimisation_problem}
Having determined a communication performance metric and a lower bound on it, MR-CaTP can be formalised as the following optimisation problem
\begin{subequations}\label{eq:opt_prob}
    \begin{align}
        \bm{u^{k*}} = \argmin_{\bm{u^k}} & \quad f(\bm{u^k}) \label{eq:cost_fun} \\
        s.t. \quad & \bm{p^{k+1}} = \bm{p^k} + \bm{u^k} \label{eq:kin_const}\\
        & ||\bm{u_i^k}||_p \leq \bar{u}, \ \forall \ i \in \mathcal{V} \label{eq:vel_const}\\
        & \lambda_2^{k+1} \geq \underline{\lambda}_2, \label{eq:com_const}\\
        & \begin{aligned}
        ||\bm{p_i^{k+1}} - \bm{p_j^{k+1}} ||_2 \geq r_i + r_j + \varepsilon \dots \\ \forall \ i \neq j \in \mathcal{V} \times \mathcal{V},
        \end{aligned} \label{eq:col_const}
    \end{align}
\end{subequations}
where \cref{eq:cost_fun} is a cost function of the robot inputs to minimise, \cref{eq:kin_const} is the robot kinematics, \cref{eq:vel_const} is a series of constraints on the norm of the inputs, \cref{eq:com_const} is a constraint on the future Fiedler value, and \cref{eq:col_const} is a series of constraints on the relative distance between robots for ensuring collision avoidance, where $\varepsilon$ is the minimum clearance. This follows the general optimisation structure for communication-aware trajectory planning (CaTP) seen in \cite{Licea2022WhenTutorial}. The communication constraint in \cref{eq:com_const} requires solving the eigenvalue problem which can be computationally demanding, and the collision avoidance constraints in \cref{eq:col_const} are non-convex quadratic constraints which are also computationally demanding to solve. In the following sections, tractable methods for estimating the Fiedler value and ensuring collision avoidance, is presented.
\section{Fiedler Value Approximation}
To constrain the Fiedler value, a method for predicting it based on the robot inputs need to be found. This can be done by predicting the Laplacian at some time $\Delta t$ into the future and calculating the Fiedler value. Another approach is to use perturbation theory, where it is estimated how much the Fiedler value is perturbed when the robots move. If the elements of the graph Laplacian $\mathbf{L^k}$ is perturbed by $\mathbf{\Delta L^k}$ the new Fiedler value can be estimated as
\begin{equation}\label{eq:Fiedler_prediction_1}
    \hat{\lambda}_2^{k+1} = \lambda_2^k + \underbrace{\bm{v_2^{k\top}}\mathbf{\Delta L^k}\bm{v_2^k}}_{\Delta\lambda_2^k},
\end{equation}
where the first term is the current Fiedler value, calculated by solution of the eigenvalue problem for the current graph Laplacian, and the second term is a mapping of the Laplacian perturbation to the Fiedler value perturbation $\Delta\lambda_2^k$. To estimate the future Fiedler value from the robot inputs, the perturbation of the graph Laplacian from the robot inputs need to be calculated. This can be done using a first-order Taylor expansion
\begin{equation}
    \mathbf{\Delta L^k} = \sum_{i=1}^N \sum_{r=1}^n \frac{d\mathbf{L}}{dp_{i,r}}\bigg\rvert_{p_{i,r}=p_{i,r}^{k}}\left(p_{i,r}^{k+1} - p_{i,r}^{k}\right)
\end{equation}
where $p_{i,r}$ denotes the $r^{th}$ element of robot $i$'s position vector $\bm{p_i}$.
Multiplying the eigenvector into the Taylor expansion gets
\begin{equation}
    \begin{aligned}
        \bm{v_2^{k\top}}\mathbf{\Delta L^k}\bm{v_2^{k}} = \sum_{i=1}^N \sum_{r=1}^n \bm{v_2^{k\top}}\frac{d\mathbf{L}}{dp_{i,r}}\bigg\rvert_{p_{i,r}=p_{i,r}^k}\bm{v_2^{k\top}}\dots \\\ \left(p_{i,r}^{k+1} - p_{i,r}^k\right)
    \end{aligned}
\end{equation}
This can be calculated as the dot product between two vectors, and using \cref{eq:discrete_model} allows \cref{eq:Fiedler_prediction_1} to be rewritten as
\begin{equation}\label{eq:Fiedler_prediction_2}
    \hat{\lambda}_2^{k+1} = \lambda_2^k + \bm{m^{k\top}}\bm{u^k},
\end{equation}
where
\begin{equation}
    \begin{split}
        \bm{m^{k}} = \left( \left( \bm{v_2^{k\top}}\frac{d\mathbf{L}}{dp_{1,1}}\bigg\rvert_{p_{1,1}=p_{1,1}^k}\bm{v_2^k},\dots, \right. \right. \\ \left. \bm{v_2^{k\top}}\frac{d\mathbf{L}}{dp_{1,n}}\bigg\rvert_{p_{1,n}=p_{1,n}^k}\bm{v_2^k} \right), \dots \\, \left( \bm{v_2^{k\top}}\frac{d\mathbf{L}}{dp_{N,1}}\bigg\rvert_{p_{N,1}=p_{N,1}^k}\bm{v_2^k},\dots, \right. \\ \left. \left. \bm{v_2^{k\top}}\frac{d\mathbf{L}}{dp_{N,n}}\bigg\rvert_{p_{N,n}=p_{N,n}^k}\bm{v_2^k} \right) \right) \in \R^{Nn}.
    \end{split}
\end{equation}
The derivative of the Laplacian with regard to $p_{i,r}$ is equal to
\begin{equation}
    \frac{d\mathbf{L}}{dp_{i,r}} = \frac{d\mathbf{D}}{dp_{i,r}} - \frac{d\mathbf{A}}{dp_{i,r}},
\end{equation}
where the derivative of each element of the adjacency matrix and the derivative of the degree matrix can be found as
\begin{equation}
    \frac{da_{ii}}{dp_{i,r}} = 0, \ \forall \ i \in \mathcal{V}, \ r \in \N_n,
\end{equation}
\begin{equation}
    \begin{split}
    \frac{da_{ij}}{dp_{i,r}}  = \frac{da_{ji}}{dp_{i,r}} = -\alpha(1 - a_{ij}) a_{ij} \frac{p_{i,r} - p_{j,r}}{||\bm{p_{i}} - \bm{p_{j}}||_2}, \dots \\ \forall \ i \neq j \in \mathcal{V} \times \mathcal{V}, \ r \in \N_n,
    \end{split}
\end{equation}
\begin{equation}
    \frac{da_{jh}}{dp_{i,r}} = 0, \ \forall \ j \neq i,h \neq i \in \mathcal{V}\times\mathcal{V}, \ r \in \N_n,
\end{equation}
\begin{equation}
    \frac{d\mathbf{D}}{dp_{i,r}} = \text{diag}\left(\frac{d\mathbf{A}}{dp_{i,r}} \bm{1_{N\times1}} \right),
\end{equation}
where $\bm{1_{N\times1}}$ is a vector of $N$ ones.
The estimated future Fiedler value in \cref{eq:Fiedler_prediction_2} can then be used to approximate the communication constraint in \cref{eq:com_const} as
\begin{equation}\label{eq:com_const_appr}
    \lambda_2^k + \bm{m^{k\top}}\bm{u^k} \geq \underline{\lambda}_2.
\end{equation}

\section{Collision Avoidance}
\begin{figure}
    \centering
    \begin{tikzpicture}[scale=0.75]
        \pgfmathsetseed{3} 
        \def\pts{}
        \xintFor* #1 in {\xintSeq {1}{5}} \do{
          \pgfmathsetmacro{\ptx}{.9*\maxxy*rand} 
          \pgfmathsetmacro{\pty}{.9*\maxxy*rand} 
          \edef\pts{\pts, (\ptx,\pty)} 
        }
        
        \xintForpair #1#2 in \pts \do{
          \edef\pta{#1,#2}
          \begin{scope}
            \xintForpair \#3#4 in \pts \do{
              \edef\ptb{#3,#4}
              \ifx\pta\ptb\relax 
                \tikzstyle{myclip}=[];
              \else
                \tikzstyle{myclip}=[clip];
              \fi;
              \path[myclip] (#3,#4) to[half plane] (#1,#2);
            }
            \clip (-\maxxy,-\maxxy) rectangle (\maxxy,\maxxy); 
            \pgfmathsetmacro{\randhue}{rnd}
            \definecolor{randcolor}{hsb}{\randhue,.5,1}
            \fill[randcolor] (#1,#2) circle (4*\biglen); 
            \fill[draw=black,fill=black,very thick] (#1,#2) circle (1.5pt); 
          \end{scope}
        }
        \pgfresetboundingbox
        \draw[white] (-\maxxy,-\maxxy) rectangle (\maxxy,\maxxy);
    \end{tikzpicture}
    \caption{Voronoi partition for 5 points in $\R^2$.}
    \label{fig:voronoi}
\end{figure}
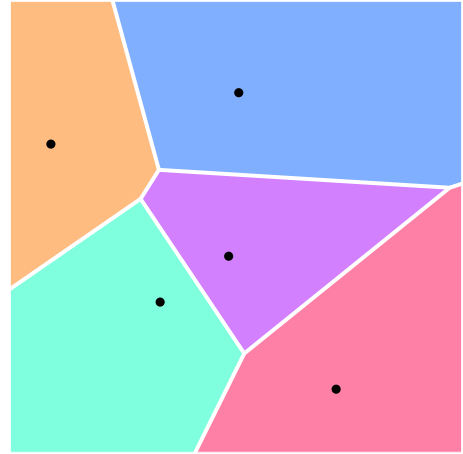

Besides constraining the inputs to the robots to ensure communication performance, they must also be constrained to ensure collision avoidance. There are a number of ways to achieve collision avoidance \cite{Khatib1985Real-timeRobots,VanBerg2008ReciprocalNavigation,VanDenBerg2011ReciprocalAvoidance}. One way is by constraining the future position of the robots to lie within disjoint sets
\begin{equation}
    \begin{gathered}
        \bm{p_i^{k+1}} \in \mathcal{P}^k_i \ \forall \ i \in \mathcal{V},\dots \\
        \mathcal{P}^k_i \cap \mathcal{P}^k_j = \emptyset \ \forall \ i\neq j \in \mathcal{V}\times\mathcal{V}.
    \end{gathered}
\end{equation}
Like in \cite{Zhou2017FastCells}, these disjoint sets are calculated as buffered Voronoi partitions. A Voronoi partition is given as
\begin{equation}
    \begin{split}
    \mathcal{P}^k_i = \{ \bm{\rho} \in \R^n \ | \ ||\bm{\rho} - \bm{p_i^k}||_2 \leq ||\bm{\rho} - \bm{p_j^k}||_2 \dots \\ \forall \ i \neq j \in \mathcal{V} \times \mathcal{V} \},
    \end{split}
\end{equation}
i.e., the partition that belongs to robot $i$ is all the points in $\R^n$ that is closer to it than all other robots, see \cref{fig:voronoi}.
One property of Voronoi partitions is that it can be represented using half-spaces
\begin{equation}
    \mathcal{P}^k_i = \{ \bm{\rho} \in \R^n \ | \ \mathbf{C_i^k}\bm{\rho} \leq \bm{d_i^k} \},
\end{equation}
which can be used as linear inequality constraints in the optimisation problem. To calculate the inequality constraint, the fact that the dual graph of the Voronoi tessellation is the Delaunay triangulation is used. The Delaunay triangulation is a graph
\begin{equation}
    \mathcal{D}(\bm{p^k}) = (\mathcal{V_D},\mathcal{E_D}), \quad \mathcal{E_D} \in \mathcal{V_D}\times\mathcal{V_D}.
\end{equation}
where the edge $(i,j) \in \mathcal{E_D}$ indicates that the Voronoi partitions of robot $i$ and $j$ are bordering each other. A Delaunay triangulation for a set of points can be found using off-the-shelf libraries, e.g., SciPy for Python \cite{Virtanen2020SciPyPython}. From this, the hyperplane separating the Voronoi partitions of robot $i$ and $j$ can be found as
\begin{equation}
    \bm{c_{ij}^{k\top}}\bm{p_i^k} \leq d_{ij}^k \ \forall \ (i,j) \in \mathcal{E_D},
\end{equation}
where $\bm{c_{ij}}$ is the normal of the border between the two partitions and $d_{ij}$ is the bias of the border from the origin
\begin{equation}
    \bm{c_{ij}^k} = \frac{\bm{p_j^k} - \bm{p_i^k}}{|| \bm{p_j^k} - \bm{p_i^k} ||_2},
\end{equation}
\begin{equation}
    d_{ij}^k = \frac{1}{2}\bm{c_{ij}^{k\top}}(\bm{p_i^k} + \bm{p_j^k}) - (r_i + \varepsilon/2). 
\end{equation}
Since the robots are not point-masses, the Voronoi partition of each robot is deflated by the radius $r_i$ of the robot and half the minimum clearance $\varepsilon$. By concatenating the transpose of the $\bm{c_{ij}^k}$ vectors together and the $d_{ij}^k$ values together for all the neighbours one can get the linear inequality constraint $\mathbf{C_i^k}\bm{p_i^k} \leq \bm{d_i^k}$ representing the Voronoi partition of robot $i$. The constraint set for the concatenated position vector $\bm{p}$ can then be found by making a block diagonal matrix and row-wise concatenation of the inequalities for all the robots
\begin{equation}
    \mathbf{C^k} = \begin{bmatrix}
        \mathbf{C_1^k} & \mathbf{0} & \dots & \mathbf{0} \\
        \mathbf{0 }& \mathbf{C_2^k} & \dots & \mathbf{0}\\
        \vdots & \vdots & \ddots & \vdots \\
        \mathbf{0} & \mathbf{0} & \dots & \mathbf{C_N^k}
    \end{bmatrix}, \quad
\end{equation}
\begin{equation}
    \bm{d^k} = (\bm{d_1^k},\bm{d_2^k},\dots,\bm{d_N^k}).
\end{equation}
The Voronoi partition can then be used to approximate the collision avoidance constraint in \cref{eq:col_const} as
\begin{equation}\label{eq:col_avoid_appr}
    \mathcal{P}^k = \{\bm{\rho}\in\R^{nN} \ | \ \mathbf{C^k}\bm{\rho} \leq \bm{d^k} \}.
\end{equation}

\section{K-step prediction}
One issue with constraining the inputs to the robots by predicting one step into the future is that it can lead to greedy solutions that do not consider how the problem will evolve. To mitigate this, the problem is transformed into a receding horizon problem where the positions of the robots and the Fiedler value is predicted $M$ steps into the future, the first input is applied, and the process is repeated indefinitely or until the algorithm is terminated. The $M$ future positions and Fiedler values are predicted as
\begin{equation}\label{eq:pos_pred}
    \bm{P^{k+1}} = \bm{1_{K\times1}}\otimes\bm{p^k} + \mathbf{B} \bm{U^k}
\end{equation}
\begin{equation}
    \bm{\hat{\Lambda}_2^{k+1}} = \lambda_2^k\bm{1_{K\times1}} + \mathbf{M^k}\bm{U^k},
\end{equation}
\begin{equation}
    \mathbf{M^k} = \mathbf{L_K} \otimes \bm{m^{k\top}}, \quad \mathbf{B} = \mathbf{L_K} \otimes \mathbf{I_{nN}}
\end{equation}
where $\mathbf{L_K}$ is a $K\times K$ lower triangular matrix of ones, $\mathbf{I_{nN}}$ is an $nN\times nN$ identity matrix, $\bm{1_{K\times1}}$ is a vector with $K$ ones, $\otimes$ is the Kronecker product operator, and
\begin{equation}
    \bm{\hat{\Lambda}_2^{k+1}} = (\hat{\lambda}_2^{k+1},\hat{\lambda}_2^{k+2},\dots,\hat{\lambda}_2^{k+K}),
\end{equation}
\begin{equation}
    \bm{P^{k+1}} = (\bm{p^{k+1}},\bm{p^{k+2}},\dots,\bm{p^{k+K}}),
\end{equation}
\begin{equation}
    \bm{U^k} = (\bm{u^k},\bm{u^{k+1}},\dots,\bm{u^{k+K-1}}).
\end{equation}
The Voronoi-based collision avoidance constraint is also altered to fit within the $K$ step prediction by applying the linear inequality constraint in \cref{eq:col_avoid_appr} to each of the $K$ predicted positions
\begin{equation}
    (\mathbf{I_K}\otimes\mathbf{C^k}) \bm{P^{k+1}} \leq \bm{1_{K\times1}}\otimes\bm{d^k},
\end{equation}
By substituting the predicted positions with \cref{eq:pos_pred}, the inequality constraint becomes
\begin{equation}
    (\mathbf{I_K}\otimes\mathbf{C^k})(\bm{1_{K\times1}}\otimes\bm{p^k} +  \mathbf{B} \bm{U^k}) \leq \bm{1_{K\times1}}\otimes\bm{d^k},
\end{equation}
\begin{equation}
    \underbrace{(\mathbf{I_K}\otimes\mathbf{C^k})\mathbf{B}}_{\mathbf{\Tilde{C}^k}}\bm{U^k} \leq \underbrace{\bm{1_{K\times1}}\otimes(\bm{d^k} - \mathbf{C^k}\bm{p^k})}_{\bm{\Tilde{d}^k}}.
\end{equation}
Thereby, the $K$ predicted positions of the robots remain within each of their Voronoi partitions generated at time step $k$, ensuring collision avoidance. The approximated constraints in \cref{eq:com_const_appr} and \cref{eq:col_avoid_appr} using an $K$-step prediction then becomes
\begin{equation}
    \mathbf{M^{k}} \bm{U^k} \geq (\underline{\lambda}_2 - \lambda_2^k)\bm{1_{K\times1}},
\end{equation}
\begin{equation}
    \mathbf{\Tilde{C}^k}\bm{U^k} \leq \bm{\Tilde{d}^k}.
\end{equation}

\section{Optimisation Problem Approximation}\label{sec:prob_approx}
Having approximated the Fiedler value $K$ steps into the future and determined a linear collisions avoidance constraint, an approximation to the optimisation problem in \cref{eq:opt_prob} can be written as
\begin{subequations}\label{eq:approx_opt_prob}
    \begin{align}
        \bm{{U^{k}}^{*}} = \argmin_{\bm{U^k}}& \ \sum_{h=0}^{K-1} f(\bm{u^{k+h}})\\ 
        s.t. \quad
        & \bm{P^{k+1}} = \bm{1_{M\times1}}\otimes\bm{p^k} +  \mathbf{B} \bm{U^k} \\
        & ||\bm{u_i^{k+h}}||_p \leq \Bar{u}, \ \forall \ i \in \mathcal{V}, \dots \\ 
        & \quad h\in\{0,\dots,K-1\}, \\
        & \mathbf{M^{k}} \bm{U^k} \geq (\underline{\lambda}_2 - \lambda_2^k)\bm{1_{K\times1}}, \label{eq:lin_com_const}\\
        &\mathbf{\Tilde{C}^k}\bm{U^k} \leq \bm{\Tilde{d}^k}. \label{eq:lin_col_const}
    \end{align}
\end{subequations}
where \cref{eq:lin_com_const} and \cref{eq:lin_col_const} are the linearised communication and collision avoidance constraints. Since the Fiedler value is predicted directly from the inputs to the robots and not from the predicted positions, the kinematic constraint in \cref{eq:kin_const} can be removed. However, it has been included in the problem approximation for consistency.

\section{Applications}
To demonstrate the efficacy of the communication constraint, two applications are presented: an inspection task, and a communication insurance service.
\subsection{Inspection Task}
In an inspection task $L$ points of interest (POI) have to be visited, where the position of the $j$\textsuperscript{th} POI is denoted as $\bm{\gamma_j} \in \R^n$. Robot $1$ is assigned as a base-station (BS) located at $\bm{p_1} = \bm{0}$. It is assumed that there are fewer POIs than available robots, i.e., $L < N$, and that each robot starts at an initial position $\bm{p_i^0}$ such that $\lambda_2 \geq \underline{\lambda}_2$. Before deploying the robots, each POI needs to be assigned to a unique robot. The problem of assigning POIs to robots can be solved as a minimum perfect matching, where the matching is perfect in the sense that every POI is assigned to a robot and each robot has at most one POI assigned to it, and minimum in the sense that the assignment has the lowest total cost of each POI being visited by the robot assigned to it. The cost for each robot-POI pair is represented as a matrix $\mathbf{\Delta} = (\bm{\delta_{ij}})\in\R_{\geq0}^{L\times N}$ where the element at the $ij$\textsuperscript{th} index is the distance between the $i$\textsuperscript{th} POI and the initial position of robot $j$, $\bm{\delta_{ij}} = || \bm{\gamma_{i}} - \bm{p_j^0} ||_2 \ \forall \ (i,j) \in \{1,\dots,L\}\times\{2,\dots,N\}$, and $\bm{\delta_{i1}} = \infty \ \forall \ i \in \{1,\dots,L\}$ since the BS cannot be assigned to a POI. The minimum matching can then be found as
\begin{equation}
    \begin{aligned}
        \mathbf{S} =& \argmin_{S\in\{0,1\}^{L\times N}} \bm{1_{L\times1}^\top}(\mathbf{S}\odot\mathbf{\Delta})\bm{1_{N\times1}}, \\
        s.t.& \quad \mathbf{S}\bm{1_{N\times1}} = \bm{1_{L\times1}}, \\
        &\quad \mathbf{S^\top}\bm{1_{L\times1}} = \bm{1_{N\times1}},
    \end{aligned}
\end{equation}
where $\odot$ is the Hadamard operator, and $\mathbf{S}$ is the matching matrix, with a $1$ at the $i,j$\textsuperscript{th} index indicating that POI $i$ has been assigned to robot $j$. This problem is solved using the Hungarian algorithm \cite{Kuhn1955TheProblem}. The K future positions of the robots assigned to the POIs can then be found as
\begin{equation}
    \bm{P_S^{k+1}} = (\mathbf{I_K}\otimes(\mathbf{S}\otimes\mathbf{I_{2}}))\bm{P^{k+1}},
\end{equation}
which, when inserting the K-step prediction model in \cref{eq:pos_pred}, can be written out as
\begin{equation}
    \begin{aligned}
    \bm{P_S^{k+1}} =& (\mathbf{I_K}\otimes(\mathbf{S}\otimes\mathbf{I_{2}}))(\bm{1_{K\times1}}\otimes\bm{p^k} + \mathbf{B}\bm{U^k}) \\
    =& \underbrace{(\mathbf{I_K}\otimes(\mathbf{S}\otimes\mathbf{I_{2}}))(\bm{1_{K\times1}}\otimes\bm{p^k})}_{\bm{P_S^k}} \dots \\ 
    &+ \underbrace{(\mathbf{I_K}\otimes(\mathbf{S}\otimes\mathbf{I_{2}})) \mathbf{B}}_{\mathbf{\Tilde{S}}}\bm{U^k}.
    \end{aligned}
\end{equation}
The error between the future positions of the robots assigned to the POIs and the POIs can be found as
\begin{equation}
    \bm{E^{k+1}_S} = \underbrace{\bm{\Gamma} - \bm{P_S^k}}_{\bm{E_S^k}} - \mathbf{\Tilde{S}}\bm{U^k}
\end{equation}
where $\bm{\Gamma} = \bm{1_{K\times1}}\otimes(\bm{\gamma_1},\dots,\bm{\gamma_L})$. The squared error is then found as
\begin{equation}
    \begin{aligned}
        \frac{1}{2}\bm{E_S^{k+1\top}}\bm{E_S^{k+1}} =& \frac{1}{2}(\bm{E_S^k} - \mathbf{\Tilde{S}}\bm{U^k})^\top(\bm{E_S^k} - \mathbf{\Tilde{S}}\bm{U^k}) \\
        =& \frac{1}{2}\bm{E_S^{k\top}}\bm{E_S^k} + \frac{1}{2}\bm{U^{k\top}}\mathbf{\Tilde{S}^\top}\mathbf{\Tilde{S}}\bm{U^k} \dots \\
        &- \bm{E_S^{k\top}}\mathbf{\Tilde{S}}\bm{U^k}.
    \end{aligned}
\end{equation}
The first term is a constant and can thus be neglected from the cost function. The remaining two terms are used in the cost function. This only penalises the robots that are assigned to POIs. The negative gradient of the Fiedler value with regard to the unassigned robots, which function as communication relays, is added as a linear cost moving them towards positions maximising the Fiedler value,
\begin{equation}
    \begin{split}
        \bm{r} = -([\mathbf{0_{1\times (K-1)}},1]\mathbf{M^k})\odot(\bm{1_{1\times K}}\otimes\dots\\(1 - \bm{1_{1\times 2K}}(\mathbf{S}\otimes\mathbf{I_2}))).
    \end{split}
\end{equation}
Lastly, a quadratic cost $\mathbf{I_{2NK}}$ on the inputs to the robots is introduced such that the robots more easily settle. The cost function $f(\bm{U})$ then becomes
\begin{equation}
    \begin{split}
        f(\bm{U^k}) = \frac{1}{2}\bm{U^{k\top}}(\mathbf{\Tilde{S}^\top}\mathbf{\Tilde{S}} + \zeta\mathbf{I_{2NK}})\bm{U^k} \dots \\ - (\bm{E_S^{k\top}}\mathbf{\Tilde{S}} + \eta\bm{r})\bm{U^k}.
    \end{split}
\end{equation}
where $\zeta,\eta\in\R_{\geq0}$ are weights. This will drive the assigned robots towards the POIs and the remaining robots towards positions maximising the Fiedler value while also constraining it. 
\subsection{Communication Insurance Service}
Even if the multi-robot planning problem cannot be formalised as an optimisation problem, the communication constraint can still be applied. One can imagine a scenario where each robot wants to move according to some input $\bm{u^k_{ref,i}}\in\R^2$ at time-step $k$. These desired inputs are then supplied to a communication insurance service (CIS) that calculates the inputs of each robot that deviates the least from their desired inputs while ensuring communication and collision avoidance. The cost function can be constructed as the squared error between the inputs of the robots $\bm{U^k}$ and the desired inputs $\bm{U_{ref}^k}$ 
\begin{equation}\label{eq:com_service_cost}
    f(\bm{U^k}) = \frac{1}{2}\bm{U^{k\top}}\bm{U^k} - \bm{U_{ref}^{k\top}}\bm{U^k} + \bm{s^\top}\mathbf{H_s}\bm{s}
\end{equation}
where
\begin{equation}
    \bm{U_{ref}^k} = \bm{1_{K\times1}}\otimes\bm{u_{ref}^k}
\end{equation}
and $\bm{u_{ref}^k} = (\bm{u_{ref,1}^{k}},\dots,\bm{u_{ref,N}^{k}})$ is the concatenated input reference. Furthermore, a slack constraint is added to the approximate optimisation problem in \cref{eq:approx_opt_prob}
\begin{equation}
        \mathbf{\Delta \Lambda_2^{k}} \bm{U} \geq (\underset{\sim}{\smash{\lambda_2}} - \lambda_2^k)\bm{1_{K\times1}} - \bm{s}, \quad \bm{s} \geq \bm{0},
\end{equation}
with the quadratic cost on the slack variable $\bm{s}$ in \cref{eq:com_service_cost} penalising the robot inputs if the Fiedler value is below the soft lower bound $\underset{\sim}{\smash{\lambda_2}} \geq \underline{\lambda}_2$. This is useful in scenarios where communication is allowed to degrade below some threshold, but it is undesired.
\section{Simulation Results}
The communication constrained inspection planner and communication insurance service is tested in simulation with the robots moving in $\R^2$ and the $\infty$-norm of the robot inputs being constrained, turning the optimisation problems into linearly constrained quadratic programs. The remaining simulation parameters can be seen in the appendix in \cref{tab:sim_param}.
\subsection{Inspection Task}
\begin{figure}[ht]
    \centering
    \includegraphics[width=\linewidth]{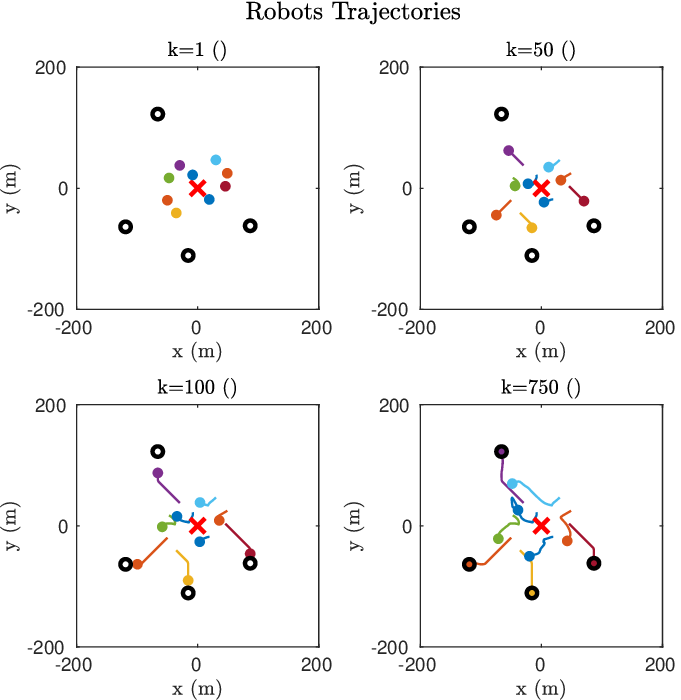}
    \caption{Positions of robots from simulation of inspection task where $N=10$ robots are deployed to inspect $L=4$ inspection points. The red cross indicates the position of the base-station.}
    \label{fig:poi_pos}
\end{figure}

\begin{figure}[ht]
    \centering
    \includegraphics[width=\linewidth]{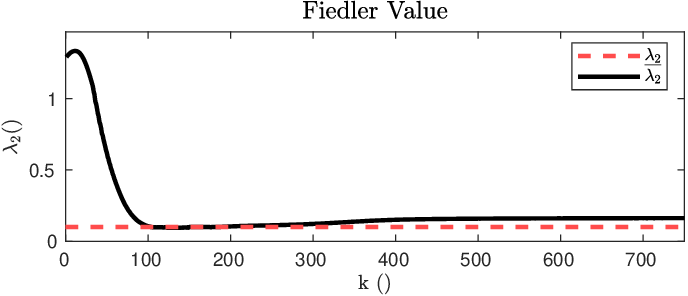}
    \caption{Fiedler over time from simulation of inspection task where $N=10$ robots are deployed to inspect $L=4$ inspection points.}
    \label{fig:poi_fiedler}
\end{figure}

\begin{figure}[ht]
    \centering
    \includegraphics[width=\linewidth]{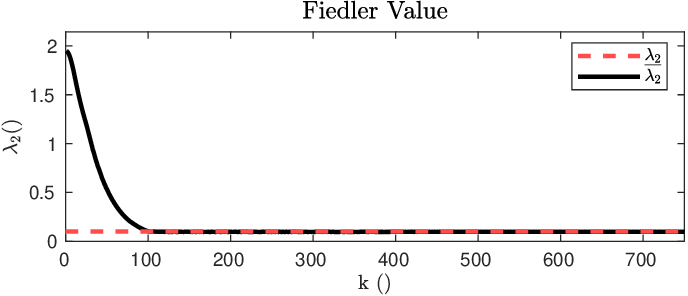}
    \caption{Fiedler over time from simulation of an infeasible inspection task where $N=10$ robots are deployed to inspect $L=4$ inspection points.}
    \label{fig:poi_fiedler_infs}
\end{figure}

In the inspection task simulation, 10 robots starting at random positions are deployed to 4 randomly sampled POIs. As can be seen on \cref{fig:poi_pos} the robots assigned to the POIs successfully reach the POIs, while the relay robot move in order to ensure communication.
As can be seen on \cref{fig:poi_fiedler}, when the robots are deployed the Fiedler value rapidly decreases until it reaches the lower bound at approximately $k=100$ iterations. Despite having reached the lower bound, the inspection robots continue moving while being supported by the relay robots, until the POIs have been reached at approximately $k=180$ iterations. However, the relay robots continue moving, increasing the Fiedler value until it settles when the relay robots have reached their locally optimal positions.
In some cases, it might not be feasible for the POIs to be inspected given the number of available robots. In such scenarios, the method will halt the robots at the positions closest to the inspection points, see \cref{fig:poi_fiedler_infs}. As can be seen in \cref{fig:poi_fiedler_infs}, the Fiedler value decreases until it has reached the lower bound where it remains.
\begin{figure}[ht]
    \centering
    \includegraphics[width=\linewidth]{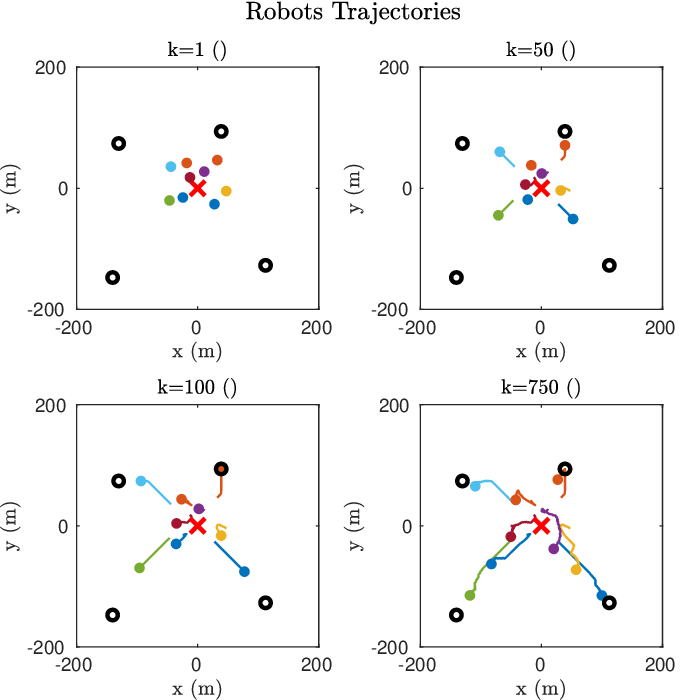}
    \caption{Positions of robots from simulation of an infeasible inspection task where $N=10$ robots are deployed to inspect $L=4$ inspection points. The red cross indicates the position of the base-station.}
    \label{fig:poi_pos_infs}
\end{figure}
To verify the justification for using a sub-optimal approximation of the original MR-CaTP problem, the optimisation problem is solved with the original constraints in \cref{eq:com_const} and \cref{eq:col_const} using Matlab's nonlinear solver fmincon.
\begin{figure}[ht]
    \centering
    \includegraphics[width=\linewidth]{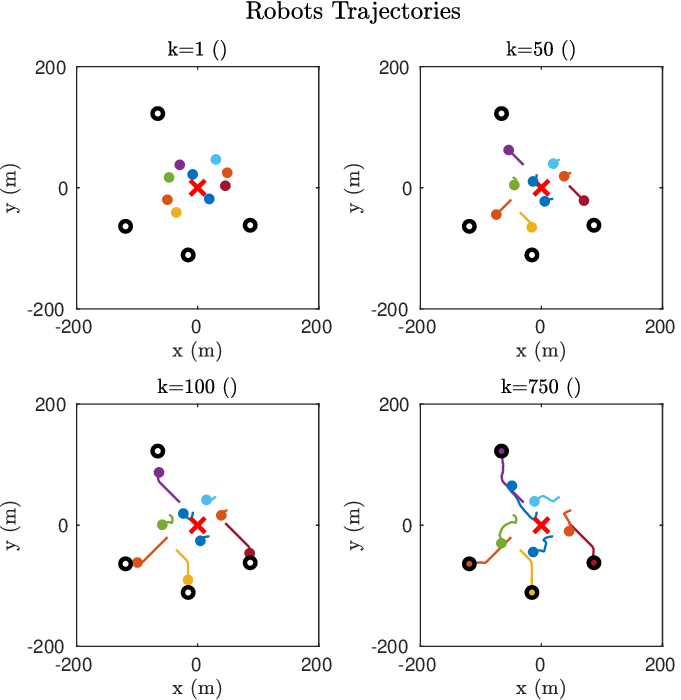}
    \caption{Positions of robots from simulation of inspection task where $N=10$ robots are deployed to inspect $L=4$ inspection points, solved using original communication and collision avoidance constraints. The red cross indicates the position of the base-station.}
    \label{fig:poi_exp}
\end{figure}
In \cref{fig:poi_exp}, the positions of the robots using the original constraints can be seen. Comparing with the positions using the approximate constraints in \cref{fig:poi_pos}, it can be seen that they are equivalent.
\begin{table}[ht]
    \centering
    \caption{Statistics of running time of each iteration in milliseconds for original and approximate problem, and factor between the two}
    \begin{tabular}{c|c|c|c|c|c}
                          & \textbf{min} & \textbf{median} & \textbf{mean} & \textbf{max} & \textbf{var} \\
        \hline \hline
        orig. prob. & 1711 & 3963 & 3823 & 5180 & 249 \\
        \hline
        approx. prob. & 2.154 & 2.883 & 3.269 & 77.1 & 
        1.21e-2  \\
        \hline
        factor & 794.3 & 1375 & 1169 & 67.19 & 20.58e3
    \end{tabular}
    \label{tab:runtime}
\end{table}
The running time of each iteration of constructing the optimisation problem and solving it is measured when using the original and approximate constraints. As can be seen in \cref{tab:runtime}, the original problem is significantly slower to solve. On average, the original problem is 3 orders of magnitude slower to solve than the approximate problem, and also has a substantially larger variance in the time to solve.
\subsection{Communication Insurance Service}
\begin{figure}[ht]
    \centering
    \includegraphics[width=\linewidth]{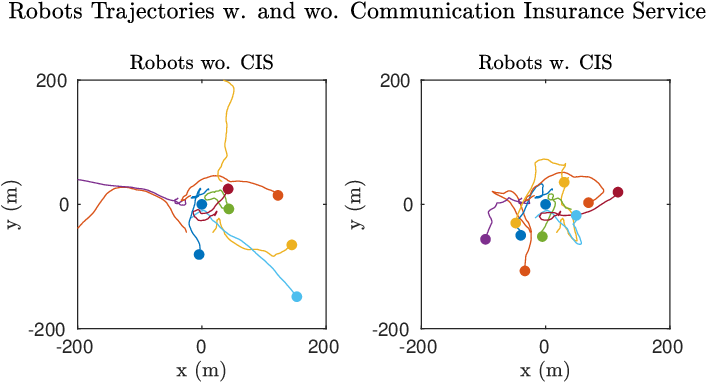}
    \caption{Positions from simulation of $N=10$ robots moving according to a stochastic process without and with communication insurance service.}
    \label{fig:pos_w_wo_com_serv}
\end{figure}

\begin{figure}[ht]
    \centering
    \includegraphics[width=\linewidth]{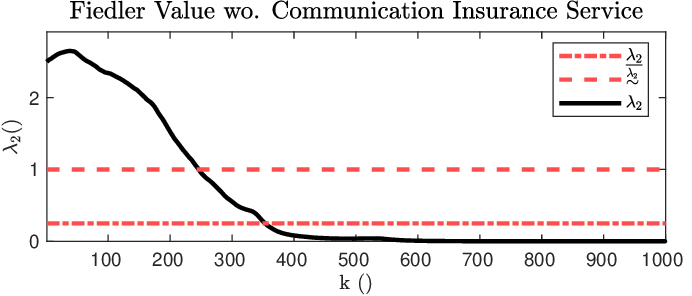}
    \caption{Fiedler value over time from simulation of $N=10$ robots moving according to a stochastic process without communication insurance service.}
    \label{fig:fiedler_wo_com_ser}
\end{figure}

\begin{figure}[ht]
    \centering
    \includegraphics[width=\linewidth]{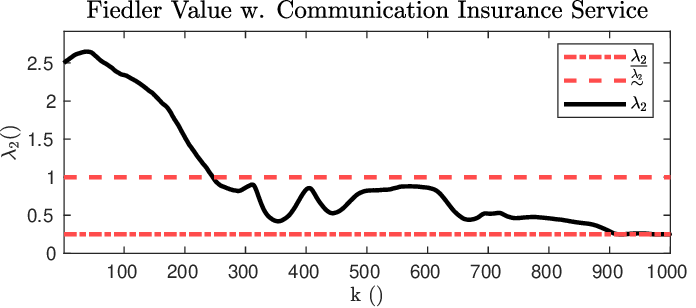}
    \caption{Fiedler value over time from simulation of $N=10$ robots moving according to a stochastic process with communication insurance service.}
    \label{fig:fiedler_w_com_ser}
\end{figure}

To test the CIS, the desired inputs of the robots follow a stochastic process where
\begin{equation}
    \begin{split}
        \bm{u_{ref,i}^k} = \bm{u_{i}^{k-1}} + \bm{\upsilon_{i}^k}, \quad \bm{\upsilon_{i}^k} \sim \mathcal{N}(\bm{0},\mathbf{\Sigma_{\bm{\upsilon}}}) \dots \\ \forall \ i \in \mathcal{V}\setminus\{1\},
    \end{split}
\end{equation}
where robot $1$ is kept stationary to serve as a base-station. As can be seen in \cref{fig:pos_w_wo_com_serv}, when the CIS is not used the robots diverge, as expected. However, when the CIS is used the robots are kept within vicinity of the base-station.
As can be seen in \cref{fig:fiedler_wo_com_ser}, when the CIS is not used the Fiedler value rapidly decreases and converges to $0$ in approximately $k=500$ iterations. However, as can be seen in \cref{fig:fiedler_w_com_ser}, when the CIS is used the Fiedler value is kept above $\underline{\lambda}_2$ and also periodically increases due to the soft constraint.
\section{Discussion}
This paper has presented a method for ensuring communication that is both tractable to solve and versatile in its applications. The method has been applied to an inspection task, and a communication insurance service. In the inspection task, one could argue that since the relay robots move to maximise the Fiedler value, the constraint is unnecessary. However, it cannot be guaranteed that the robots tasked with inspecting the POIs do not move faster than the relay robots can move in order to ensure that the Fiedler value stays above the lower bound, validating the use of a method for constraining the Fiedler value. There are some limitations in the method, which will be addressed in the following.
\subsection{Discrepancy in Fiedler value derivative}
The most obvious limitation is due to the linearisation of the Fiedler value derivative. Since the Fiedler value derivative is linearised, there will be a discrepancy between the predicted Fiedler value and the actual future Fiedler value. By keeping the bound on the change in position $\Bar{u}$ sufficiently small, this discrepancy will most likely be negligible. However, as no bound on the discrepancy and no guarantees on the smoothness of the Fiedler value has been found, how small $\Bar{u}$ should be is unknown. The risk of the discrepancy is largest when the Fiedler value is close to the lower bound $\underline{\lambda}_2$, as it might go below this even though the predicted Fiedler value does not. This might be mitigated by decreasing $\Bar{u}$ as $\lambda_2$ gets closer to $\underline{\lambda}_2$, but is not something that has been investigated in this paper. When $\lambda_2$ falls below $\underline{\lambda}_2$, the system can recover as long as the gap is less than what the robots can manage to change within one planning period, given the constraints on the robot inputs. If this is not possible, the robots should move to maximise the Fiedler value until $\lambda_2 \geq \underline{\lambda}_2$.
\subsection{Robot dynamics}
In this method, the robots are assumed to be holonomic. For ground robots with omnidirectional drive and multirotors this assumption is valid. However, a large number of robots are non-holonomic, and their limitations have to either be handled explicitly by including a model of their kinematics in the constraints or implicitly by constraining the change in position. If the robots are modelled as linear time-invariant (LTI) systems, the dynamics can be accounted for directly in the formulation by substituting the robot inputs in the constraints with the dynamics of the robots. Lastly, this method also assumes that the robots are able to halt if needed. For some robots, e.g., gliders, this is not possible, and a long prediction horizon might be needed such that feasible communication preserving manoeuvres can be calculated.
\subsection{Locally optimal solutions}
As the method is limited to predict $K$ steps into the future, only locally optimal solutions are found. As such, the solutions are highly dependent on the starting positions of the robots. To avoid local minima, a higher level communication-aware planner might be needed. E.g., for the inspection task, a planner, such as the one in \cite{Shi2021Communication-AwareMaximization}, could calculate the optimal terminal positions of the robots, but not consider how to ensure communication while the robots transition from their starting positions to their optimal positions. This could then be ensured by the planner presented in this paper.
\subsection{Versatility}
One of the claimed benefits of the method is its versatility. Since the method does not consider the objective of the robot, it can be applied to any problem that can be formalised as an optimisation problem on the change in robot position. Furthermore, it is demonstrated that even in the case where the problem cannot be formalised as an optimisation problem, the method can be applied as a communication insurance service that finds the smallest deviations to the desired inputs ensuring communication and collision avoidance. It is assumed that most multi-robot objectives can be met by controlling the change in position of the robots. This assumption is deemed acceptable since position control is integrated directly into the control stack of most robots, and therefore the CIS can be injected into the control stack.
\subsection{Number of deployed robots}
In this paper it is assumed that all $N$ robots are deployed. However, there might be scenarios where only a subset of the robots need to be deployed to fulfil an objective, rendering any additionally deployed robot redundant. One could a priori calculate how many robots are needed to solve a given task and only deploy those, however, it can be difficult or even impossible to calculate how many robots are needed for a given task. This makes a dynamic deployment scheme, where robots are deployed as needed, pertinent.
\subsection{Input constraint}
As can be seen in \cref{fig:poi_pos}, the robots tend to move diagonally. This is due to the $\infty$-norm constraint on the inputs to the robots, making diagonal movements more favourable. This can be mitigated by using another norm, e.g., the $2$-norm. This is not a linear constraint and can therefore not be solved by linearly constrained quadratic programming solvers, but it is convex and can therefore be solved using, e.g., CVX \cite{Grant2013MatlabBeta.,Grant2008GraphPrograms}.
\subsection{Infeasibility}
In some situations the robot inputs might become over-constrained, resulting in the optimiser not being able to find a solution. In this case, a strategy must be used that ensure communication until the robots have moved to a position where the optimiser can find a solution. This could be done using a communication-aware rendezvous strategy as in, e.g., \cite{Su2010RendezvousConnectivity}. In this approach, the robots will move together until the Fiedler value reaches some threshold or until the optimisation problem becomes feasible. To avoid limit-cycles, it is most likely necessary for rendezvous to continue for some time after the optimisation problem becomes feasible.
\subsection{Obstacles}
In this paper, obstacles have not been included. To allow for this, additional collision avoidance constraints and line-of-sight constraints would have to be imposed on the robot system. 
\section{Conclusion}
In this paper, the reformulation of a generic MR-CaTP problem to make it more tractable was presented. In the proposed method, the constraint on the future Fiedler value is reformulated into a linear constraint on the robot inputs. Furthermore, a linear constraint on the robot inputs to ensure collision avoidance was also presented. The efficacy of the method was presented with two use-cases: an inspection task, and a communication insurance service. As demonstrated through simulation, the constraint is able to ensure that the Fiedler value stays above a specified lower bound while the robots move to achieve their goals. Furthermore, by comparing the running time of the MR-CaTP algorithm using the approximate constraints with the original constraints, it is evident that there is a clear advantage and need for using the approximate constraints. A thorough discussion on the limitations of the method and some of their possible mitigations was also presented. In \cite{Derenick2010AEnvironments}, the authors present a similar approach to the one presented in this paper. The main difference is that in their paper, the Fiedler value is constrained using semi-definite programming. The main benefit of constraining the Fiedler value using a linear constraint is that the contribution of each robot to the Fiedler value is calculated explicitly. This opens the possibility for distributing the optimisation problem across the robots, making the problem of multi-robot communication-aware trajectory planning more scalable.
\appendix
\begin{table}[ht]
    \centering
    \caption{Simulation Parameters}
    \begin{tabular}{c|c|c}
        \textbf{Variable} & \textbf{Value} & \textbf{Unit} \\
        \hline
        \hline
        $t_s$ & $0.2$ & $s$ \\
        $\Delta t$ & $0.4$ & $s$ \\
        $N$ & $10$ & \\
        $L$ & $4$ & \\
        $d_{50}$ & $50$ & $m$ \\
        $\alpha$ & $0.1$ & $m^{-1}$ \\
        $\underline{\lambda}_2$ inspection task & $0.1$ & \\
        $\underline{\lambda}_2$ CIS & $0.25$ & \\
        $\underset{\sim}{\smash{\lambda_2}}$ CIS & $1$ & \\
        $r$ & $0.1$ & $m$ \\
        $\varepsilon$ & $10$ & $m$ \\
        $\zeta$ & $0.1$ & \\
        $\eta$ & $1\text{e}^{3}$ & \\
        $\mathbf{H_s}$ & $0.5\mathbf{I_{M}}$ & \\
        $\mathbf{\Sigma_v}$ & $0.1\mathbf{I_{2N}}$ &
    \end{tabular}
    \label{tab:sim_param}
\end{table}
\printbibliography

\end{document}